\documentclass[review]{elsarticle}

\usepackage{lineno,hyperref}
\modulolinenumbers[5]

\journal{Neurocomputing}

\usepackage[dvipsnames]{xcolor}

\usepackage{graphicx}
\usepackage{times}
\usepackage{epsfig}
\usepackage{graphicx}
\usepackage{amsmath}
\usepackage{amssymb}
\usepackage{pifont}
\usepackage{float}
\usepackage{xcolor}
\usepackage{bm}
\usepackage{times}
\usepackage{epsfig}
\usepackage{graphicx}
\usepackage{float}
\floatstyle{plaintop}
\restylefloat{table}
\usepackage{multirow}
\usepackage{array}
\newcolumntype{x}[1]{>{\centering\arraybackslash\hspace{0pt}}p{#1}}
\usepackage{subcaption}
\usepackage[symbol]{footmisc}
\usepackage{soul,color}
\usepackage{graphicx}
\usepackage{times}
\usepackage{epsfig}
\usepackage{graphicx}
\usepackage{amsmath}
\usepackage{amssymb}
\usepackage{pifont}
\usepackage{float}
\usepackage{xcolor}
\usepackage{bm}
\usepackage{times}
\usepackage{epsfig}
\usepackage{graphicx}
\usepackage{float}
\floatstyle{plaintop}
\restylefloat{table}
\usepackage{multirow}
\usepackage{array}
\newcolumntype{x}[1]{>{\centering\arraybackslash\hspace{0pt}}p{#1}}
\usepackage{subcaption}
\usepackage[symbol]{footmisc}
\usepackage{wrapfig}
\usepackage{booktabs}
\usepackage{stfloats, caption}%

\usepackage{times}
\usepackage{epsfig}
\usepackage{graphicx}
\usepackage{amsmath}
\usepackage{amssymb}
\usepackage{footnote}
\usepackage{dsfont}

\usepackage{comment}
\usepackage{blindtext}
\usepackage{morewrites}
\usepackage{makecell}
\usepackage{pifont}%
\usepackage{multirow}
\usepackage{soul}

\soulregister\cite7
\soulregister\ref7
\soulregister\pageref7

\usepackage[utf8]{inputenc} %
\usepackage[T1]{fontenc}    %
\usepackage{hyperref}       %
\usepackage{url}            %
\usepackage{booktabs}       %
\usepackage{amsfonts}       %
\usepackage{nicefrac}       %
\usepackage{microtype}      %
\usepackage{xcolor}         %

\usepackage[utf8]{inputenc} %
\usepackage[T1]{fontenc}    %
\usepackage{hyperref}       %
\usepackage{url}            %
\usepackage{booktabs}       %
\usepackage{amsfonts}       %
\usepackage{nicefrac}       %
\usepackage{microtype}      %
\usepackage{xcolor}         %

\usepackage{wrapfig}
\usepackage{graphicx}
\usepackage{amsmath}
\usepackage{mathtools}
\usepackage{multirow}
\usepackage{makecell}
\usepackage{tabularx}
\usepackage{algorithm2e}
\usepackage{fancyvrb,xcolor}
\usepackage{nicematrix}
\usepackage{bbm}

\usepackage{algorithm2e}
\usepackage{algorithmic}
\usepackage{fancyvrb}
\usepackage{xcolor}
\usepackage{nicematrix}

\usepackage{tikz}
\usepackage{comment}
\usepackage{amsmath,amssymb} %
\usepackage{color}
\usepackage{enumitem}
\usepackage{amsthm}
\newcommand{\Bac}[1]{\textcolor{black}{#1}}

\usepackage{multirow}
\usepackage{makecell}
\usepackage{amsmath}
\usepackage{capt-of}
\usepackage{tabularx}
\usepackage{epsfig}
\usepackage{amssymb}
\usepackage{amsfonts}
\usepackage{booktabs}
\usepackage{scalerel}
\usepackage{listings}
\usepackage{varwidth}
\usepackage[export]{adjustbox}
\usepackage{tikz}
\usetikzlibrary{tikzmark}

\usepackage{stmaryrd}
\usepackage{bbm}
\usepackage{wrapfig}
\usepackage{pifont}

\definecolor{deepblue}{rgb}{0,0,0.5}
\definecolor{officeblue}{RGB}{0,102,204}
\definecolor{deepred}{rgb}{0.6,0,0}
\definecolor{deepgreen}{rgb}{0,0.5,0}
\definecolor{mybrickred}{RGB}{182,50,28}

\definecolor{fillcolor}{RGB}{216,217,252}

\begin{document}

\begin{frontmatter}

\title{Brainformer: Mimic Human Visual Brain Functions to Machine Vision Models via fMRI}

\author{Xuan-Bac Nguyen$^{1}$, Xin Li$^{2}$, Pawan Sinha$^{3}$,  Samee U. Khan$^{4}$, Khoa Luu$^{1}$\\
    $^{1}$ CVIU Lab, University of Arkansas, AR 72703 \quad $^{2}$ University at Albany, NY 12222 \\
    $^{3}$ Massachusetts Institute of Technology, MA 02139 \\ $^{4}$  Mississippi State University, MS 39762 \\
	\tt\small \{xnguyen,khoaluu\}@uark.edu, xli48@albany.edu,\\
        \tt\small psinha@mit.edu, skhan@ece.msstate.edu
}

\begin{abstract}
Human perception plays a vital role in forming beliefs and understanding reality. A deeper understanding of brain functionality will lead to the development of novel deep neural networks. In this work, we introduce a novel framework named Brainformer, a straightforward yet effective Transformer-based framework, to analyze Functional Magnetic Resonance Imaging (fMRI) patterns in the human perception system from a machine-learning perspective. Specifically, we present the Multi-scale fMRI Transformer to explore brain activity patterns through fMRI signals. This architecture includes a simple yet efficient module for high-dimensional fMRI signal encoding and incorporates a novel embedding technique called 3D Voxels Embedding. Secondly, drawing inspiration from the functionality of the brain's Region of Interest, we introduce a novel loss function called Brain fMRI Guidance Loss. This loss function mimics brain activity patterns from these regions in the deep neural network using fMRI data. This work introduces a prospective approach to transferring knowledge from human perception to neural networks. Our experiments demonstrate that leveraging fMRI information allows the machine vision model to achieve results comparable to State-of-the-Art methods in various image recognition tasks.
\end{abstract}

\begin{keyword}
Self-supervised Learning, Artificial Intelligence,  Vision, Human Neuroscience, Scene Understanding, fMRI
\end{keyword}

\end{frontmatter}

\section{Introduction}
\label{sec:intro}

Recent studies in machine vision understanding \cite{resnet, densenet, krizhevsky2012imagenet, vaswani2017attention, liu2021swin, dosovitskiy2020image} have demonstrated the effectiveness of single-encoder models through pretraining on image datasets, e.g., ImageNet \cite{imagenet}. These methods are designed to acquire universal visual representations of objects that can be flexibly applied to various downstream tasks, including object detection and semantic segmentation. Nonetheless, the most significant limitation of these methods lies in the costly annotation process, mainly when applied at a large scale. In response to this challenge, self-supervised techniques have emerged \cite{bao2021beit, he2022masked, Nguyen_2023_CVPR}.
They aim to acquire visual representations without incurring human annotation expenses while delivering commendable performance compared to supervised methods. 

To avoid expensive annotation efforts, a surge in the development of foundational language models, e.g., BERT \cite{BERT}, GPT-2, GPT-3 \cite{brown2020language}, RoBERTa \cite{liu2019roberta}, T5 \cite{raffel2020exploring}, BART \cite{lewis2019bart} is observed. These approaches are typically trained on extensive datasets and use the text to guide the vision models \cite{yuan2021florence, clip, jia2021scaling}. 
From the recent success of text as supervision for visual learning, we revisit 
one of the key questions in the early days of artificial intelligence: \textit{What if human brain behaviors can serve as the guiding force for the machine vision models?}. To answer this question, we found that Functional Magnetic Resonance Imaging (fMRI) has provided valuable insights into various aspects of human cognition and neuroscience. It contains rich information on how the human visual system works. For example,  fMRI can help identify the specific regions of the brain that are active during various cognitive tasks. Studies \cite{30graves2016hybrid, 31gregor2015draw,32xu2015show} have shown that different intelligence-related tasks activate specific brain areas, such as problem-solving or memory. In addition, human intelligence is not just about the activity of individual brain regions but also about how different brain regions communicate. fMRI can reveal functional connectivity patterns, highlighting networks involved in various cognitive processes. 

While text typically reflects the outcomes of the recognition process, fMRI captures the dynamics of the cognitive processes involved. For instance, examiners are tasked with describing the context of an image, such as ``A man is sitting on the chair." Initially, they focus on the man and then shift attention to the chair, determining the interactions between these objects, specifically, the act of sitting. Subsequently, they conclude the context in the text form. When encountering this description for the first time, it remains unclear which object the examiner attended to first or whether other smaller objects influenced their perception. Notably, \textbf{this information can be encoded in fMRI signals}. By analyzing specific brain regions or regions of interest,  we can explore the whole recognition process of the human brain.
This work provides a promising approach to further studies and endeavors to bridge the gap between human intelligence and deep neural networks. 
The contributions of this work can be summarized as follows.

\begin{figure}[!t]
    \centering
    \includegraphics[width=1.0\linewidth]{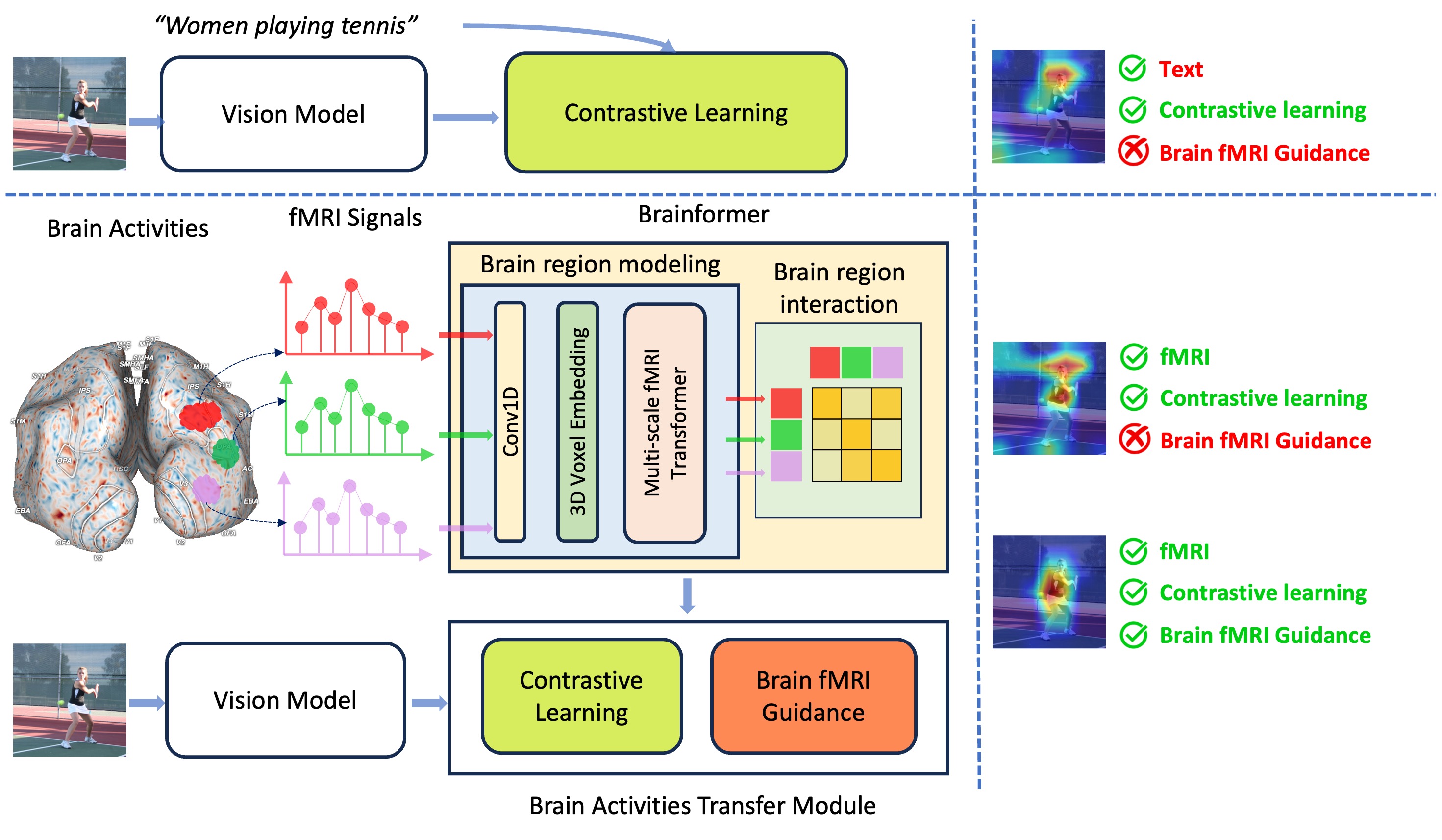}
    \caption{Given a pair of images and fMRI signals (x-axis is the voxel index, y-axis is the magnitude of the voxel response), Brainformer can explore the local patterns of fMRI signals from brain regions and discover their interactions. \textbf{Best view in color}}
    \label{fig:abstract_figure}
\end{figure}

\noindent
\textbf{Contributions of this Work.} This paper presents a new simple but efficient Transformer-based framework, Brainformer, to analyze and leverage fMRI signals captured from human brain activities to supervise the machine vision learning model. 
Second, a new Brain fMRI Guidance Loss function is presented to distill information from fMRI extracted using Brainformer. It serves as a guidance signal to enhance the performance capabilities of vision models.
Finally, Brainformer is designed for self-supervised learning and trained as an end-to-end deep network. It consistently outperforms prior State-of-the-Art Self-supervised learning methods across standard benchmarks, i.e., object detection, instance segmentation, semantic segmentation, and brain response prediction. The code will be released.

\section{Related Work}
\label{sec:related_work}

\noindent
\textbf{Vision-Language Foundation Models}. 
Pretraining on large-scale datasets \cite{imagenet, mahajan2018exploring, zhai2022scaling} has become a popular approach for many visual recognition problems \cite{he2016deep,he2022masked,nguyen2019audio,nguyen2019sketch,nguyen2021clusformer,nguyen2022multi,nguyen2022two,nguyen2023fairness,nguyen2024bractive,nguyen2024diffusion,nguyen2024qclusformer,nguyen2024hierarchical,serna2024video}, e.g., classification, localization, or segmentation, etc. Despite this method has great success, it is limited by the number of annotated training data that is costly to collect. Recently, self-supervised training has been proposed to address this problem \cite{nguyen2020self, nguyen2024insect,nguyen2023micron}. Significant advancements have occurred in vision-language pertaining \cite{radford2021learning, jia2021scaling, zhai2022lit, pham2023combined, wang2021simvlm, wang2022ofa, li2022blip, yu2022coca, zhai2023sigmoid, luo2023lexlip, wang2023equivariant}. This approach aims to encode both visual and text information into a model \cite{tan2019lxmert, chen2020uniter, zhang2021vinvl, ren2015faster, kim2021vilt, bao2022vlmo,nguyen2024insect}. 
Recent work has proposed Image-Text foundation models that bridge the gap between image and text understanding \cite{clip, align, piergiovanni2022answer, wang2021simvlm, wang2022ofa}. They leverage dual-encoder architecture pre-trained on noise image-text pairs using contrastive learning or generative losses. Their studies showcase impressive performance in vision-language benchmarks while maintaining strong visual encoder capabilities for classification tasks.
\newline
\noindent
\textbf{Leveraging Human Brain Functions In Training}. While the text has successfully demonstrated its efficiency in helping machine learning models enhance their performance, recent studies have been inspired by human brain mechanisms to improve models \cite{thomas2022self, thual2022aligning, kan2022brain, portes2022distinguishing, lin2022mind, millet2022toward, fong2018using, safarani2021towards, 10073607, pogoncheff2023explaining, khosla2022characterizing, cui2022fuzzy, sarch2023brain, duan2024few, lee2024hyper, thomas2022self, quesada2022mtneuro}. Some recent methods have used neural activity data to guide the training of models \cite{palazzo2020decoding, 34fong2018using, Spampinato2016deep, nishida2020brain, nechyba1995human}. They utilized EEG and fMRI signals to constrain the neural network to behave like the neural response in the visual cortex.

\noindent
\textbf{Decoding functional MRI}.
Decoding visual information from fMRI signals has been studied for a decade \cite{haynes2005predicting, thirion2006inverse, kamitani2005decoding, cox2003functional, haxby2001distributed}. Most of these studies aimed to explore the hidden information inside the brain. It is a difficult task because of the low signal-to-noise ratio. Recently, with the help of deep learning, the authors in \cite{chen2023seeing, scotti2023reconstructing, takagi2023high, ozcelik2023natural, lin2022mind, ozcelik2023brain} proposed methods to reconstruct what humans see from fMRI signal using diffusion models. 
Different from recent studies on decoding fMRI \cite{takagi2023high, scotti2023reconstructing, chen2023seeing}, our goal is to explore valuable vision information from fMRI signal, i.e., semantic and structure information, and utilize them as supervision for helping to enhance the recognition capability of the vision model. Although our approach also aims to extract feature representations of fMRI data, these previous methods still have limitations and do not apply to our problem. 
First, prior approaches, such as those by Yu et al. \cite{takagi2023high}, and Paul et al., \cite{scotti2023reconstructing} relied on linear models and Multi-layer Perceptron to generate fMRI feature representations. This approach cannot capture the local pattern of the signals, especially the non-linear or complicated patterns. Kim et al. \cite{kim2023swift} presented SwiFT for self-training fMRI in 4D data, which does not apply to our problem. On the other hand, MindVis \cite{chen2023seeing} utilized a Masked Brain Model inspired by MAE, which explored local patterns but ignored the correlations between multiple voxels, which describes how neurons interact inside the brain. Furthermore, MindVis failed to leverage the 3D spatial information in fMRI signals, which is crucial in understanding the relationships between voxels. Most notably, none of these approaches leverage characteristics of functional regions of interest (ROI) within the brain, which contain rich information about visual stimuli.

\section{The Proposed Approach}
\label{sec:method}

\subsection{Motivation}
\label{sec:motivation}
The fMRI contains a structured and semantically rich representation of information since fMRI captures brain activities in the context of a person engaging in visual tasks. It provides ground truth data about how the human brain responds to visual stimuli. According to recent studies \cite{press2001visual, kanwisher2001faces, dwivedi2021unveiling}, the brain can be divided into several Regions of Interest (ROI), where each region holds a different function. \Bac{
The ROIs corresponding to different perceived objects are determined by brain functional localizers (floc). For instance, consider the \textit{floc-face} example. A set of visual stimuli is prepared to determine these regions, alternating between categories such as faces, objects, scenes, and words. fMRI signals are recorded from multiple participants as they are exposed to these stimuli. To pinpoint the \textit{floc-faces} regions, the brain's response to faces is contrasted with its responses to other categories (e.g., faces vs. objects) using statistical methods like General Linear Models (GLM)}.
Especially in the scope of visual cognition, these are 6 regions specified as Early retinotopic visual regions (\textit{prf-visualrois}), Body-selective regions (\textit{floc-bodies}), Face-selective regions (\textit{floc-faces}), Place-selective regions (\textit{floc-places}), Word-selective regions (\textit{floc-words}) and Anatomical streams (\textit{streams}). 
In general, these regions are the responses for processing information related to the human body, face, place, motion, or objects regarding identification, recognition, etc. Therefore, the fMRI signals from these regions can be used to train computer vision models, providing a reference for what the brain is ``seeing" during the training. By training vision models jointly with brain activities observed in fMRI scans, they can explore the relationship between the visual information processed by the brain and the visual information that models are learning. It can help uncover how the brain represents and processes visual information.

\subsection{Brainformer}

\begin{figure*}[!ht]
    \centering
    \includegraphics[width=1.0\columnwidth]{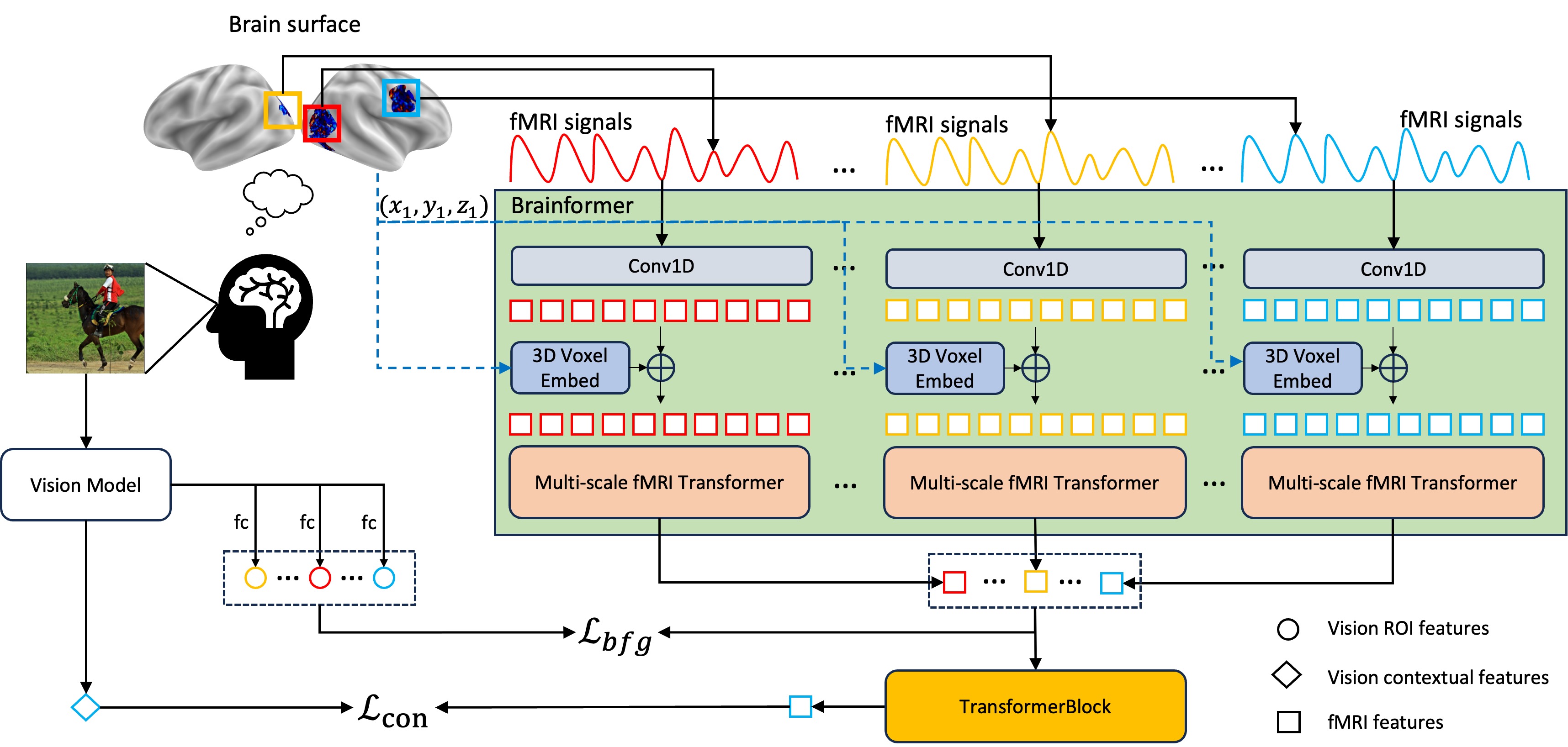}
    \caption{Brainformer utilizes fMRI signals (x-axis is the voxel index, y-axis is the magnitude of the voxel response) from specific brain regions as input, extracting the local features representing patterns within each region. The $\texttt{TransformerBlock}$ measures the correlation among these regions to emulate brain activities. This information is subsequently transferred to the vision model through Contrastive Loss and Brain fMRI Guidance Loss.}
    \label{fig:brainformer}
\end{figure*}

\begin{figure*}[!b]
    \centering
    \includegraphics[width=1.0\columnwidth]{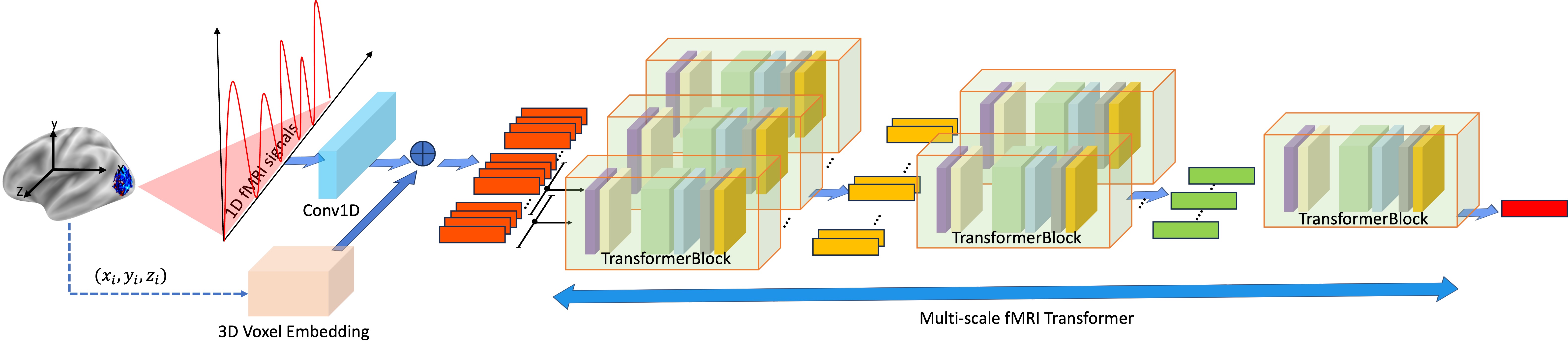}
    \caption{The details of Multi-scale fMRI Transformer module.}
    \label{fig:multi_scale}
\end{figure*}

As aforementioned, human intelligence is presented by the activity of individual brain regions of interest and how they interact with each other. Inspired by this concept, we introduce Brainformer, as shown in Figure \ref{fig:brainformer}, a model that takes fMRI signals from Regions of Interest in the brain as inputs. The proposed model comprises two primary modules. First, we introduce a Mutli-scale fMRI Transformer, as shown in Figure \ref{fig:multi_scale}, designed to uncover local patterns of brain activities within each ROI. Subsequently, we feed a sequence of ROI features derived from the output of the Mutli-scale fMRI Transformer into a conventional Transformer to estimate the correlation and interaction among multiple ROIs. In the following sections, we focus on the design of the Mutli-scale fMRI Transformer, as detailed in Section \ref{sec:sht}. Section \ref{sec:brain_cognitive_features} introduces brain cognitive features which represent for correlation between brain regions. Prior to that, we present an efficient way to extract features of raw signals in Section \ref{sec:fmri_feature}. Additionally, we introduce a novel technique called Brain 3D Voxel Embedding in Section \ref{sec:voxel_embeding}, aimed to preserve the spatial information of the signals.

\subsubsection{High-dimensional fMRI Feature Encoding}
\label{sec:fmri_feature}

Let $m^k$ be the fMRI signals of the $k^{th}$ region of interest in the brain. This signal can be constructed using Eqn. \eqref{eq:fmri}.
\begin{equation}
    \label{eq:fmri}
    m^k = \left[\delta(x_i, y_i, z_i)\right]_{i=0}^{N^k - 1}
\end{equation}
where $\delta$ is the function that takes the value of change in blood and oxygenation in the voxel coordinated at $(x_i, y_i, z_i)$. $N^k$ is the number of voxels in this region. This definition shows that fMRI is a 1D high-dimension signal due to the significant value of $N^k$, e.g., $N^k \approx 20K$.
A straightforward approach to model this signal is adopting linear or fully connected layers \cite{scotti2023reconstructing} and extracting its latent embedding. This approach, however, has two drawbacks. First, fully connected layers with high dimensional features are inefficient as it is challenging to learn useful information from the input space while maintaining a high memory usage of the model. Second, it focuses more on the global structure while ignoring the local patterns presented in the fMRI signal. To solve these two problems, we propose \texttt{Conv1D} to extract features of the fMRI signal. Formally, the embedding feature of fMRI is represented in Eqn. \eqref{eqn:fmri_embedding}.
\begin{equation} \label{eqn:fmri_embedding}
\begin{split}
    \mathbf{r}^k & = \texttt{Conv1D}(m^k) \in \mathbb{R} ^ {N^k \times d_r}
\end{split}
\end{equation}
The $\mathbf{r}^k$ is the embedding features of $m^k$ after the convolution step, and $d_r$ is the embedding dimension.

\subsubsection{Brain 3D Voxel Embedding}
\label{sec:voxel_embeding}

\begin{figure}[!ht]
    \includegraphics[width=0.9\linewidth]{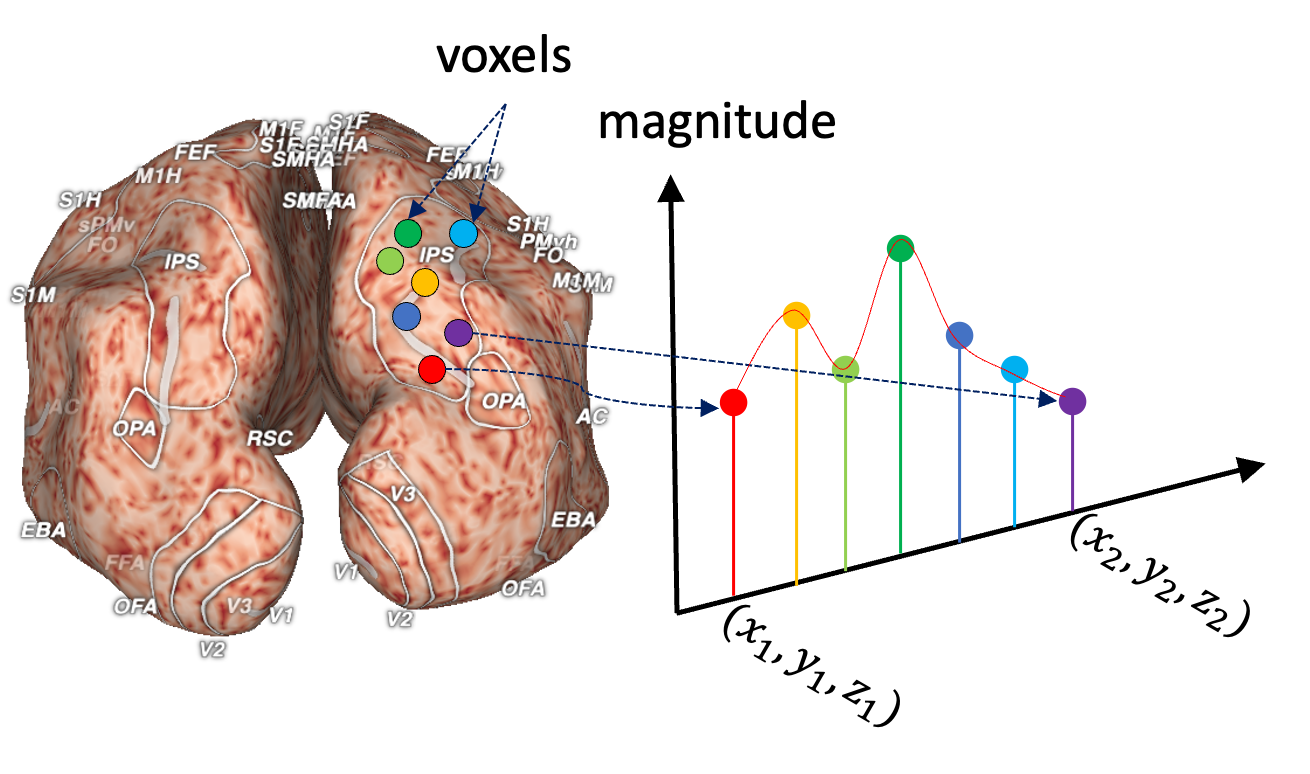}
    \caption{Two voxels, denoted $v_1 = (x_1, y_1, z_1)$ and $v_2 = (x_2, y_2, z_2)$ that are located closely in the 3D space of MRI, but in the fMRI signals (x-axis is the voxel index, y-axis is the magnitude of the voxel response), they are far awya from each other.}
    \label{fig:voxel_problem}
\end{figure}

As shown in Eqn \eqref{eq:fmri}, the fMRI signal is a compressed representation of Magnetic Resonance Imaging (MRI) containing detailed 3D information.
However, direct learning from the flattened signal will \textit{ignore the spatial structure}. Figure \ref{fig:voxel_problem} illustrates this problem. 
Consider two voxels, denoted as $v_1 = (x_1, y_1, z_1)$ and $v_2 = (x_2, y_2, z_2)$, which are closely situated in the 3D space of MRI, but in fMRI signals representation, they appear distant from each other. Therefore, it becomes crucial to incorporate spatial information for the model to uncover deeper structural features.
Meanwhile, prior studies \cite{scotti2023reconstructing, takagi2023high, chen2023seeing} still need to address this problem. In light of this concept, we introduce a new approach named \textit{Brain 3D Voxel Embedding} to capture the spatial architecture of fMRI more effectively.

Let $m_i^k$ be the voxel in the signal $m^k$ and $(x_i^k, y_i^k, z_i^k)$ are the 3D coordinates of this voxel. We use $\texttt{Linear}$ function to map these $(x_i^k, y_i^k, z_i^k)$ coordinates into the same dimension, i.e., $d_r$ dimension, as the fMRI embedding features in Eqn. \eqref{eqn:fmri_embedding} 
\begin{equation} \label{eqn:voxel_embedding}
\begin{split}
    v_i^k &= \texttt{Linear}(x_i^k, y_i^k, z_i^k) \in \mathbb{R}^{d_r} \\
    v(m^k) &= \texttt{concat}\left[v_0^k, v_1^k, \dots, v_{N^k-1}^k\right] \in \mathbb{R}^{N^k \times d_r}
\end{split}
\end{equation}
where $v_i^k$ is the brain 3D voxel embedding of single voxel $i^{th}$, $v(m^k)$ is the embedding of the entire signal $m^k$.
Incorporating with \textit{Brain 3D Voxel Embedding} to the Eqn. \eqref{eqn:fmri_embedding}, we get the features of fMRI as shown in Eqn. \eqref{eqn:fmri_voxel_embedding}. 
\begin{equation} \label{eqn:fmri_voxel_embedding}
\begin{split}
    \mathbf{r}^k &= \texttt{Conv1D}(m^k) + v(m^k)
\end{split}
\end{equation}
\subsubsection{Mutli-scale fMRI Transformer}

\label{sec:sht}

While extremely powerful for a wide range of natural language processing tasks, Transformers have limitations when handling long sequences. These limitations are primarily due to the quadratic complexity of the self-attention mechanism and the model's parameter count. In addition, the attention score might collapse and close to zero if the sequence length is considerable. 
Hence, using the typical transformer for lengthy fMRI signals 
is not a complete solution. 

To deal with the long sequence in our problem, we propose the Mutli-scale fMRI Transformer, shown in Figure \ref{fig:multi_scale}. It contains multiple levels, each level consisting of a Transformer block denoted as $\texttt{TransBlock}$. The length of the signal is reduced when passing through each level of the network until it is sufficient enough. 
The details of the Multi-scale fMRI Transformer are presented in Algorithm \ref{algo:sht}.
In particular, each level consists of  two steps:
\begin{itemize}
\item[] \textbf{Step 1:} We employ a slicing window with $w$ of width that traverses the entire signal $\mathbf{r}^k$ from its beginning to the end, with a step size of $s$. This process decomposes $\mathbf{r}^k$ into smaller sub-sequences, denoted as $\mathbf{q}_i^k = \mathbf{r}^k\left[i * s:i * s+w\right] \quad 0 \leq i \leq n_s = \frac{N^k}{s}$ that has a computation-efficiency length for the Transformer. The adjacent subsequences overlap by a distance of $s$, preventing the loss of local information, and $n_s$ is the number of subsequences.

\item[] \textbf{Step 2:} We fed $\mathbf{q}_i^k$ into the $\texttt{TransBlock}$ to learn the patterns inside the signal and get the features $\Bar{\mathbf{q}}_{i}^k = \texttt{TransBlock}^k(\mathbf{q}_i^k)$. After that,  all the features $\Bar{\mathbf{q}}_{i}^k$ of subsequences are gathered to form a new sequence
    $\textbf{t} = \left[\Bar{\mathbf{q}}_{i}^k\right]_{i=0}^{n_s-1}$ and then passed into the next level. 
\end{itemize}

\begin{algorithm}[H]
\centering
\small
\caption{Mutli-scale fMRI Transformer}
\label{algo:sht}
\begin{algorithmic}[1]
\STATE {\bfseries Input:} The feature $\mathbf{r}^k$ of $k^{th}$ fMRI signals , window size $w$, stride: $s$, number of level $h$. 
\STATE {\bfseries Output:} The fMRI features $\Bar{\mathbf{q}}^k$

\STATE{$\mathbf{x} \gets \mathbf{r}^k$}

\FOR{$j=1, \cdots, h$}
    \STATE{$\mathbf{t} \gets \text{empty list}$} 
    \STATE{$L \gets |x|$}
    \FOR{$i=0, \cdots \frac{L}{s}$}
        \STATE{$\mathbf{q}_i^k \gets \mathbf{x}\left[i * s:i * s+w\right]$}
        \STATE{$\Bar{\mathbf{q}}_{i}^k \gets \texttt{TransBlock}^k_j(\mathbf{q}_i^k)$}
        \STATE{$\mathbf{t} \gets \left[\mathbf{t} \quad \Bar{\mathbf{q}}_{i}^k\right]$}
    \ENDFOR
    \STATE{$\mathbf{x} \gets \mathbf{t}$}
\ENDFOR
\STATE{$\Bar{\mathbf{q}}^k \gets \texttt{TransBlock}(\mathbf{x})$}
\STATE{\textbf{return}  $\Bar{\mathbf{q}}^k$}
\end{algorithmic}
\end{algorithm}

\subsubsection{Brain Cognitive Features}
\label{sec:brain_cognitive_features}
Let $\left[\Bar{\mathbf{q}}^k\right]_{k=0}^{N_r}$ be the list of fMRI features that are the outputs from the Mutli-scale fMRI Transformer. $N_r$ is the number of regions in the brain. In particular, we define $N_r = 6$ as discussed in the section \ref{sec:motivation}. 
To learn the correlation between the ROIs, we adopt another feature embedding process by feeding $\left[\Bar{\mathbf{q}}^k\right]_{k=0}^{N_r}$ into a final $\texttt{TransBlock}$ and receiving the brain cognitive features, denoted as $\mathbf{q} = \texttt{TransBlock} (\left[\Bar{\mathbf{q}}^k\right]_{k=0}^{N_r})$

\subsection{Training Objectives}

Our framework is trained by a combination of Contrastive Loss and Brain fMRI Guidance Loss as $\mathcal{L} = \lambda_{con} \mathcal{L}_{con} + \lambda_{bfg} \mathcal{L}_{bfg}$
where $\lambda_{con}$ and $\lambda_{bfg}$ are the weights of the loss functions. In this paper, we select $\lambda_{con} = \lambda_{bfg} = 0.5$. The details of $\mathcal{L}_{con}$ and $\mathcal{L}_{bfg}$ are described in the sections below.
\subsubsection{Contrastive Loss}

Let $\mathbf{p} \in \mathbb{R}^{d_r}$ be the image features extracted by the image encoder, e.g., SwinTransformer, ConvNext. We employ contrastive loss \cite{chen2020simple} to align visual representation with the brain cognitive features $\mathbf{q}$ as in Eqn. \eqref{eqn:contrastive}.
\begin{equation} \label{eqn:contrastive}
\begin{split}
    \mathcal{L}_{con} &= -\frac{1}{N}\sum_i^N \log\frac{\exp(\mathbf{p}_i \otimes \mathbf{q}_i/\sigma)}{\sum_j^N\exp(\mathbf{p}_i \otimes \mathbf{q}_j/\sigma)} - \frac{1}{N}\sum_i^N \log\frac{\exp(\mathbf{q}_i \otimes \mathbf{p}_i/\sigma)}{\sum_j^N\exp(\mathbf{q}_i \otimes \mathbf{p}_j/\sigma)}
\end{split}
\end{equation}
where $\sigma$ is the learnable temperature factor,
$N$ is the number of samples, and $\otimes$ is the dot product. 

\subsubsection{Brain fMRI Guidance Loss}
\label{sec:guidance_loss}

Besides the contrastive loss that aligns the $\textit{global context}$ of brain signals and stimuli image, the Brain fMRI Guidance Loss aims to align the $\textit{local context}$. Since the ROI features $\Bar{\mathbf{q}}^k$ is extracted from a particular region of the brain, e.g., floc-bodies, floc-faces, etc., it embeds the information on how the brain perceives the objects inside the image.
If we can leverage these features as the guidance during training, 
the vision model can mimic the perception system of the human.

\begin{figure}
    \centering
    \includegraphics[width=0.9\linewidth]{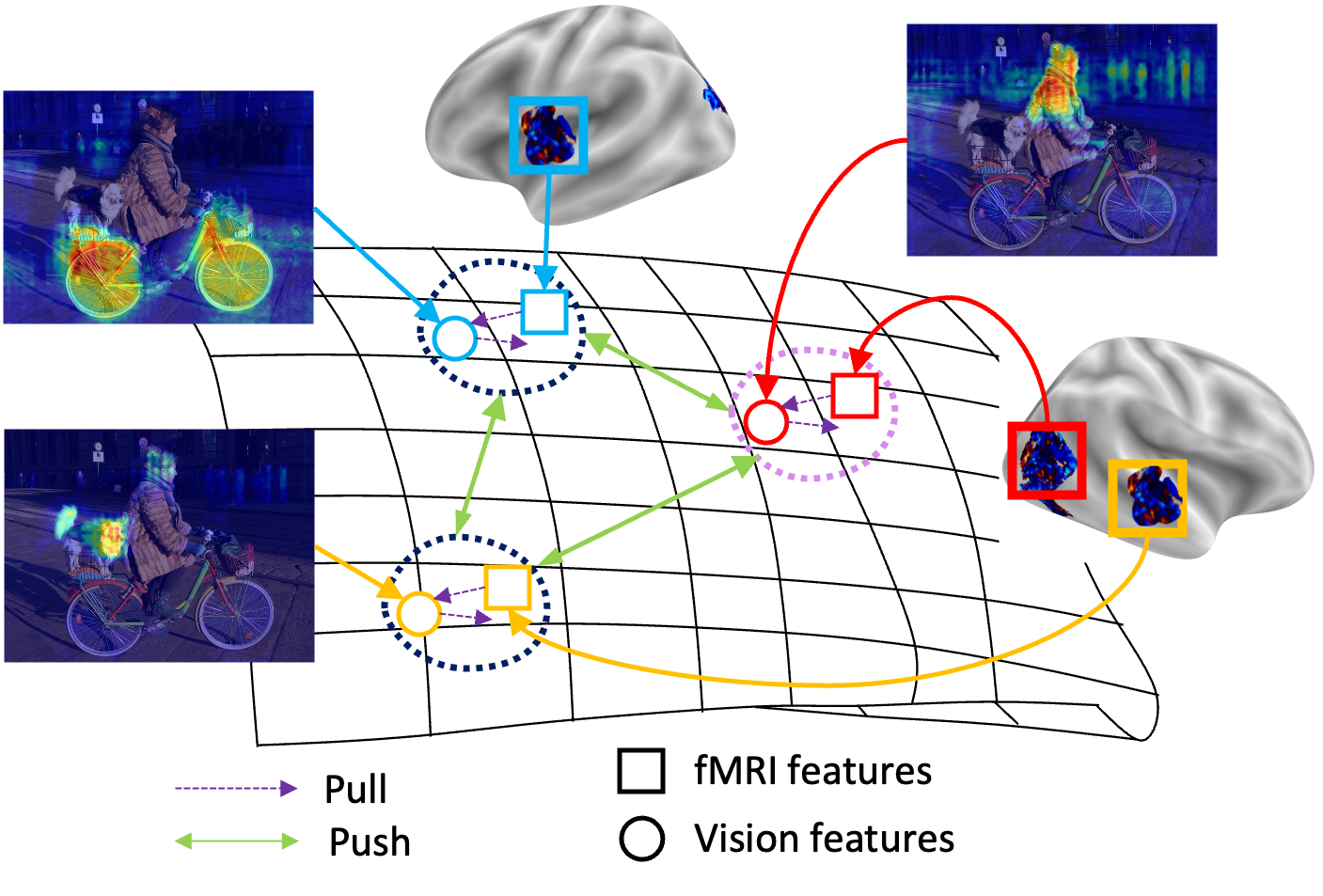}
    \caption{The circle and rectangle represent vision and fMRI features, respectively. Each color indicates a different object of interest that the human brain is processing. The Brain fMRI Guidance Loss aims to align visual and fMRI features of the same object while discriminating with features of other objects.}
    \label{fig:bs_loss}
\end{figure}

Let $\Bar{\mathbf{p}}^k$ be the vision features representing a specific local context within the image, e.g., objects, persons, the background, or locations. These features can be achieved by passing pooling features of the vision model to the linear layer and then projected to the same dimension space as $\Bar{\mathbf{q}}^k$. Our goal is to encourage the visual perceptions of the vision model to be similar to the ROI features of the brain. This requires maximizing the similarity between two features, $\Bar{\mathbf{p}}^k$ and $\Bar{\mathbf{q}}^k$. Furthermore, it is important to emphasize that the brain's ROIs have distinct functions. Consequently, we must ensure the dissimilarity of $\Bar{\mathbf{q}}^k$ with respect to all $\Bar{\mathbf{q}}^{g}$, where $k \neq g$. To facilitate the two constraints, we propose the Brain fMRI Guidance Loss as in Figure \ref{fig:bs_loss} and in Eqn. \eqref{eqn:guidance}.
\begin{equation} \label{eqn:guidance}
\begin{split}
    \mathcal{L}_{bfg} &= - \frac{1}{N} \sum_i^N \sum_k^{N_r} \log\frac{\exp(\Bar{\mathbf{p}}^k_i \otimes \Bar{\mathbf{q}}^k_i)}{\sum_g^{N_r}\exp(\Bar{\mathbf{p}}^k_i \otimes \Bar{\mathbf{q}}^g_i)}
\end{split}
\end{equation}

\section{Experimental Results}
\label{sec:results}
\subsection{Datasets}

To pretrain Brainformer, we leverage the Natural Scenes Dataset (NSD) \cite{allen2022massive}, a comprehensive compilation of responses from eight subjects obtained through high-quality 7T fMRI scans.
Each subject was exposed to approximately 73,000 natural scenes, forming the basis for constructing visual brain encoding models.

\subsection{Brainformer Training}
\Bac{We use data from seven subjects in NSD for training and leave one for testing}. 

\noindent
\textbf{\Bac{Brainformer configurations.}} \Bac{For the fMRI signals, we divide them into six regions of interest following \cite{algonauts} and feed them into Brainformer simultaneously. Experimentally, we select $\texttt{Conv1D}$ with kernel size and stride of 32 and 16, respectively. The Multi-scale fMRI Transformer is designed with a window size as $w = 64$, stride as $s = 32$, and the number of levels of the Multi-scale fMRI Transformer is set as $h=2$. Feature dimensions of image and fMRI signals are projected into $d_r = 768$ space. The 3D Voxel Embedding module is designed by a linear layer that projects 3D coordinates $(x, y, z)$ to a $d_r = 768$ dimension space. These 3D coordinates are available to use in the NSD database}.

\noindent
\textbf{\Bac{Vision models.}} \Bac{The images are resized to 224 $\times$ 224 before feeding into the vision model. We do not use any augmentations for the image because the fMRI signals depend on the input image. Any change in vision stimuli will affect human recognition. In this work, we select, but not limited to, Swin-S \cite{liu2021swin} and ConvNext-S \cite{liu2022convnet} as the image backbone}.

\noindent
\textbf{\Bac{Training strategy}.} \Bac{Brainformer is easily implemented in the Pytorch framework and trained on 8 $\times$ A100 GPUs. The initial learning rate is 0.0001 and decreases gradually following the ConsineLinear policy \cite{cosine}. A batch size of 64 per GPU is employed. Optimization is performed using AdamW \cite{adamw} with 100 epochs, with the entire training process concluding within two hours. The pre-trained models are used for further downstream tasks, including COCO object detection \cite{coco} and ADE20K semantic segmentation \cite{ade20k}}.
\begin{table}[!h]
\centering
    \caption{Results of object detection and instance segmentation on COCO dataset}
    \label{tab:coco_object_detection}
    \begin{tabular}{c|c|ccc}
    \Xhline{1.0pt}
    \multicolumn{5}{c}{{(a) Object Detection}}  \\  
    \hline
    Backbone & Pretrain & AP$^\text{box}$ & AP$^\text{box}_\text{50}$ & AP$^\text{box}_\text{75}$ \\
    \hline
     Swin-S &  Random init & {41.3($\pm$0.2)} & {63.4($\pm$0.3)} & {45.5($\pm$0.4)} \\
     Swin-S &  CLIP \cite{clip} & {41.9($\pm$0.1)} & {64.0($\pm$0.4)} & {46.0($\pm$0.3)} \\
     Swin-S &  Brainformer & \textbf{43.6($\pm$0.2)} & \textbf{65.8($\pm$0.5)} & \textbf{47.4($\pm$0.3)} \\
     \hline
     ConvNext-S &  Random init & {42.6($\pm$0.3)} & {65.5($\pm$0.2)} & {47.0($\pm$0.2)}\\
     ConvNext-S &  CLIP \cite{clip} & {42.8($\pm$0.1)} & {66.1($\pm$0.2)} & {48.3($\pm$0.3)} \\
     ConvNext-S &  Brainformer & \textbf{45.1($\pm$0.1)} & \textbf{68.2($\pm$0.2)} & \textbf{50.0($\pm$0.2)} \\               
    \hline
    \multicolumn{5}{c}{{(b) Semantic Segmentation}}  \\
    \hline
     Backbone & Pretrain & AP$^\text{segm}$ & AP$^\text{segm}_\text{50}$ & AP$^\text{segm}_\text{75}$ \\
    \hline
     Swin-S &  Random init  & {38.4($\pm$0.2)} & {60.8($\pm$0.3)} & {41.3($\pm$0.2)}\\
     Swin-S &  CLIP \cite{clip} & {39.2($\pm$0.3)}  & {61.3($\pm$0.3)} & {41.7($\pm$0.2)}\\
     Swin-S &  Brainformer  & \textbf{43.1($\pm$0.2)} & \textbf{63.1($\pm$0.5)} & \textbf{43.3($\pm$0.2)} \\
     \hline
     ConvNext-S &  Random init  & {39.8($\pm$0.4)} & {61.7($\pm$0.1)} & {42.9($\pm$0.3)}\\
     ConvNext-S &  CLIP \cite{clip} & {40.8($\pm$0.4)}  & {62.2($\pm$0.2)} & {43.5($\pm$0.1)} \\
     ConvNext-S &  Brainformer & \textbf{44.0($\pm$0.2)} & \textbf{64.0($\pm$0.5)} & \textbf{45.3($\pm$0.4)} \\   
     \hline
\end{tabular}
\end{table}

\subsection{Object Detection on COCO}
\textbf{Settings}. We conducted object detection and instance segmentation experiments using the COCO 2017 dataset, which comprises 118,000 training images, 5,000 validation images, and 20,000 test-dev images. Notably, we excluded images from the NSD dataset to prevent any data leakage, a subset of COCO. For these experiments, we utilized the MaskRCNN framework \cite{maskrcnn} implemented with mmdetect \cite{mmdetection} for enhanced performance efficiency. 

\noindent
\textbf{Performance}. The object detection and instance segmentation results are presented in Table \ref{tab:coco_object_detection}. In summary, our approach, which leverages fMRI signals for training, consistently outperforms the CLIP framework, where models rely solely on textual 
supervision. Specifically, when comparing Swin-S/CLIP to Swin-S/Random, the Swin-S/CLIP achieves approximately a 0.6\% improvement in box/AP and a 0.8\% improvement in seg/AP. However, our method Swin-S/Brainformer presents even better performance, surpassing Swin-S/CLIP by approximately 1.7\% and 3.9\% in object detection and instance segmentation, respectively, translating to a substantial 2.3\% and 4.7\% improvement compared to the same model Swin-S/Random without any pretraining method.
We also observe the same results with the ConvNext-S backbone, where Brainformer achieves around 2.3\% and 3.2\% higher than CLIP in box/AP and seg/AP.

\subsection{Semantic Segmentation on ADE20K}
\textbf{Settings}. For semantic segmentation, we conducted training using UpperNet \cite{uppernet} on the ADE20K database, which includes a wide spectrum of 150 semantic categories. This dataset comprises 25,000 images, with 20,000 allocated for training, 2,000 for validation, and 3,000 for testing.
\newline
\noindent
\textbf{Performance}. Table \ref{tab:semantic_segmentation} presents the semantic segmentation results. From these results, Swin-S/Brainformer is $1.48\%$ mIoU (41.77 v.s 40.29) higher than  Swin-S/CLIP while ConvNext-S/Brainformer maintains the performance better than ConvNext-S/CLIP by $1.65\%$ mIoU.
\begin{table}[!t]
    \centering
    \caption{Results of semantic segmentation on ADE20K and brain activities response prediction on NSD.}
    \label{tab:semantic_segmentation}
    \begin{tabular}{c|c|c|c}
        \Xhline{1.0pt}
        & & \multicolumn{1}{c|}{ADE20K} & \multicolumn{1}{c}{NSD}\\
        \multicolumn{1}{c|}{Backbone} & \multicolumn{1}{c|}{Pretrain} & mIoU & PCC\\
        \hline
        Swin-S & Random init & 38.37($\pm$0.2) & 40.41($\pm$0.4) \\
        Swin-S & CLIP \cite{clip} & 40.29($\pm$0.3) & 41.25($\pm$0.3) \\
        Swin-S & Brainformer & \textbf{41.77($\pm$0.3)} & \textbf{44.63($\pm$0.2)} \\
        \hline
        ConvNext-S & Random init & 39.22($\pm$0.3) & 54.21($\pm$0.3) \\
        ConvNext-S & CLIP \cite{clip} & 41.27($\pm$0.4) & 55.70($\pm$0.2) \\
        ConvNext-S & Brainformer & \textbf{42.92($\pm$0.3)} & \textbf{57.43($\pm$0.4)}\\
        \hline
    \end{tabular}
\end{table}

\vspace{-6mm}
\subsection{Human Brain Response Prediction NSD}
\textbf{Settings}. This experiment aims to predict the human brain response to complex natural scenes, as recorded during participants' observations \cite{algonauts, nsd}. The dataset contains responses from eight subjects in the NSD. We use seven subjects for pretraining CLIP and Brainformer, while one is reserved for the downstream task. We follow the evaluation protocols outlined in \cite{algonauts}, utilizing the Pearson Correlation Coefficient (PCC) score as our evaluation metric.\\
\noindent
\textbf{Performance}. The results of our brain response prediction are shown in Table \ref{tab:semantic_segmentation}. Significantly, Brainformer yields substantial performance improvements in this task. Specifically, Swin-S/Brainformer demonstrates approximately 4.22\% and 3.38\% better performance than Swin-S/Random and Swin-S/CLIP, respectively. Furthermore, it is worth noting that ConvNext-based models perform strong predictive capabilities, with ConvNext-S/Brainformer achieving a PCC of 57.43\%, approximately 1.73\% higher than ConvNext-S/CLIP.

\section{Ablation Studies}

\begin{table}[]
\centering
\addtolength{\tabcolsep}{-2.5pt}
\caption{Performance on various settings.}
\label{tab:ablation}
\begin{tabular}{c|cc|c|c}
\Xhline{1.0pt}
 & \multicolumn{2}{c|}{COCO} & \multicolumn{1}{c|}{ADE20K} & \multicolumn{1}{c}{NSD} \\
& AP$^\text{box}$ & AP$^\text{segm}$ & mIoU & PCC \\
\hline

pos embed & 41.9 & 41.8 & 41.72 & 56.19 \\
3D voxel embed &  \textbf{45.1} & \textbf{44.0} & \textbf{42.92} & \textbf{57.43} \\
\hline
w/o Brain fMRI Guidance Loss & 42.7 & 40.7 & 41.80 & 56.25 \\
w/ Brain fMRI Guidance Loss & \textbf{45.1} & \textbf{44.0} & \textbf{42.92} & \textbf{57.43} \\
\hline
$w = 128, s = 64$  & 41.9 & 41.1 & 40.90 & 55.80 \\
$w = 128, s = 32$  & 42.3 & 41.6 & 41.80 & 56.20 \\
$w = 64, s = 32$  & \textbf{45.1} & \textbf{44.0} & \textbf{42.92} & \textbf{57.43} \\
\hline
$\#$ subjects = 1  & 42.5 & 39.3 & 39.15 & 54.20 \\
$\#$ subjects = 3  & 42.9 & 39.6 & 39.37 & 54.51 \\
$\#$ subjects = 5  & 43.4 & 40.4 & 40.08 & 55.01 \\
$\#$ subjects = 7  & \textbf{45.1} & \textbf{44.0} & \textbf{42.92} & \textbf{57.43} \\
\hline
\Bac{fMRI Feature Encoding (Linear) }  & 42.8 & 39.1 & 39.54 & 54.39 \\
\Bac{fMRI Feature Encoding (ours) } & \textbf{45.1} & \textbf{44.0} & \textbf{42.92} & 
\textbf{57.43} \\
\hline
\Bac{$\lambda_{con} = 0.3, \lambda_{bfg} = 0.7$}  & 45.0 & 43.7 & 42.57 & \textbf{57.84} \\
\Bac{$\lambda_{con} = 0.5, \lambda_{bfg} = 0.5$} & \textbf{45.1} & \textbf{44.0} & \textbf{42.92} & {57.43}\\
\Bac{$\lambda_{con} = 0.7, \lambda_{bfg} = 0.3$}  & 45.0 & 43.8 & 39.54 & 57.20 \\
\Xhline{1.0pt}
\end{tabular}
\end{table}

\begin{figure*}[!t]
    \centering
    \includegraphics[width=0.99\linewidth]{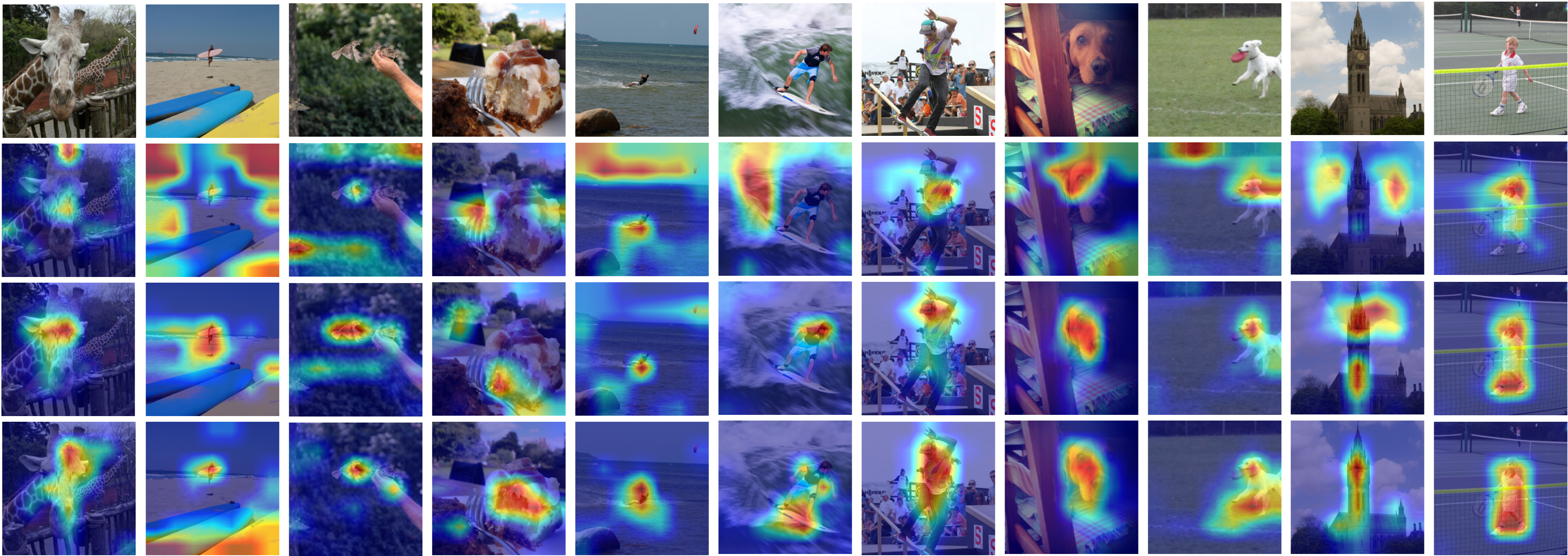}
    \caption{Visual attention with respect to fMRI signals. 
    The first row is the input images. The second row is the corresponding attention maps of the vision model training with CLIP. The third row is the results of training with Brainformer without {Brain fMRI Guidance Loss}. The last row is the results with Brainformer and guidance of {Brain fMRI Guidance Loss}. The warmer colors, the higher attention. \textbf{Best view in color}.}
    \label{fig:frmi_attention}
\end{figure*}

In this section, we study the efficiency of the Brain 3D Voxel Embedding as presented in Section \ref{sec:voxel_embeding}, Brain fMRI Guidance Loss in Section \ref{sec:guidance_loss}, hyper-parameters of Multi-scale fMRI Transformer in Section \ref{sec:sht}, and performance of Brainformer on different amounts of data. We select ConvNext-S as the backbone for these ablation studies. 
\newline
\noindent
\textbf{Brain 3D Voxel Embedding}.
In Table \ref{tab:ablation}, we provide an ablation study for 3D voxel embedding. The use of this embedding yields an improvement of approximately +3.2\% box/AP, +2.2\% seg/AP, +1.2\% mIoI, and +1.24\%PCC over the network employing traditional positional embedding in various tasks such as object detection, instance segmentation, semantic segmentation, and brain response prediction, respectively. These findings demonstrate the effectiveness of the 3D voxel embedding for fMRI signals.
\newline
\noindent
\textbf{Brain fMRI Guidance Loss}.
We also provide an ablation study for the usage of Brain fMRI Guidance Loss in Table \ref{tab:ablation}. The model trained with the guidance loss performs better than not using it. In particular, this loss function helps to improve +2.4\% box/AP, +3.3\% seg/AP, +1.12\% mIoU, and 1.18\%PCC for object detection, instance segmentation, semantic segmentation, and brain response prediction, respectively. We also investigate how Brain fMRI Guidance Loss facilitates the transfer of semantic and structured information from fMRI signals to the vision model. To accomplish this, we follow a three-step process. First, we extract the fMRI features and image features from Brainformer and the vision model, respectively. Second, we compute the cosine similarity between these feature vectors. Finally, we leverage GradCam, as described in \cite{grad_cam}, by analyzing the backward gradient of the similarity to generate an attention map. This map highlights specific image regions strongly correlated to the fMRI signals. We visualize this attention map in Figure \ref{fig:frmi_attention}.
It illustrates the effectiveness of the Brain fMRI Guidance Loss in transferring brain activities from the fMRI signals to the vision model.
\Bac{Notably, we observe that the model trained with CLIP often produces attention maps that are largely out of focus with respect to the objects in the image (second row). In contrast, Brainformer demonstrates improved focus on the objects; however, its attention maps tend to cover only parts of the objects rather than capturing the full texture (third row). Particularly, Brainformer trained with the Brain fMRI Guidance Loss generates significantly better attention maps, effectively focusing on entire objects with greater precision.}
\newline
\noindent
\textbf{Hyper-parameters for Multi-scale fMRI Transformer}
We studied the effectiveness of the hyperparameters of Brainformer, i.e., window size and stride, on the overall performance.
As shown in Table \ref{tab:ablation}, the results highlight that smaller window sizes and strides lead to better performance. Increasing the window size from 64 to 128 resulted in a 2.8\% decrease in box/AP, a 2.4\% decrease in seg/AP, a 1.12\% decrease in mIoU, and a 1.23\% decrease in PCC for object detection, instance segmentation, semantic segmentation, and brain response prediction, respectively. This is attributed to the fact that a smaller window size allows Brainformer to capture more local information. Similarly, we observed similar results when keeping the window size constant and increasing the stride from 32 to 64, with larger strides causing the model to miss more information potentially.
\newline
\noindent
\textbf{Performance on different amounts of data}.
\label{sec:abl_amount_data}
We investigate the performance of Brainformer with respect to the amount of training data.
We train the Brainformer with a different number of subjects included in the pre-training step. The performance is reported in Table \ref{tab:ablation}. It is clear that when we use only one subject data, the performance is not much improved compared to the random initialization. However, when we increased the number of subjects, the performance of Brainformer also increased accordingly. With respect to 7 subjects' training data, the performance is boosted by +2.6\% for box/AP, +4.7\% for seg/AP, +3.77\% for mIoU, and 3.23\% for PCC.
\newline
\noindent
\Bac{\textbf{Compare to existing fMRI encoding methods.} We compare the effectiveness of our proposed fMRI encoding strategy to the recent approaches. Following the previous studies \cite{takagi2023high, mindeye1, psychometry, mindbridge, mindeye2, neuropictor}, we employ a simple linear layer to extract the feature of fMRI signals from the ROIs and then use their features to guide the vision model. The performance is reported in  Table \ref{tab:ablation}. Our proposed fMRI encoding strategy clearly outperforms the existing method, achieving improvements of +2.3\% in box/AP, +4.9\% in seg/AP, +3.38\% in mIoU, and +3.04\% in PCC. This is because of the limitations of the linear layer in capturing voxel correlations and the absence of spatial information in the baseline approach. In contrast, our method effectively addresses these limitations.}
\newline
\noindent
\Bac{\textbf{Effectiveness of loss weights.} 
In this section, we evaluate the impact of the loss weights, $\lambda_{con}$ and $\lambda_{bfg}$, on performance. Specifically, we experiment with different pairs of loss weights: ($\lambda_{con} = 0.3, \lambda_{bfg} = 0.7$), ($\lambda_{con} = 0.5, \lambda_{bfg} = 0.5$), and ($\lambda_{con} = 0.7, \lambda_{bfg} = 0.3$). The results are summarized in Table \ref{tab:ablation}.
First, we observe that the PCC metric achieves the highest score with the setting ($\lambda_{con} = 0.3, \lambda_{bfg} = 0.7$). In contrast, box/AP, seg/AP, and mIoU attain their best performance when using the balanced setting ($\lambda_{con} = 0.5, \lambda_{bfg} = 0.5$). These results can be explained by the different focus of the two loss components: the Brain fMRI Guidance Loss emphasizes local information, such as facial and bodily details, while the contrastive loss emphasizes the global context. Consequently, the Human Brain Response Prediction task (evaluated using PCC) benefits more from a higher $\lambda_{bfg}$, reflecting the importance of local information.
Striking a balance between global and local information is crucial. Thus, we select equal loss weights for the contrastive loss and Brain fMRI Guidance Loss, i.e., ($\lambda_{con} = 0.5, \lambda_{bfg} = 0.5$).}
\newline
\noindent
\Bac{\textbf{Computational Complexity.} We analyze the network size and computational complexity under various settings, with results summarized in Table \ref{tab:abl-complexity}. Notably, two key parameters, i.e., window size ($w$) and stride ($s$) as defined in Section \ref{sec:sht}, play a crucial role in computational complexity. Specifically, when $w = 128$, reducing the stride $s$ from 64 to 32 nearly doubles the Floating-Point Operations Per Second (FLOPs), from 51.86G to 102.92G, while improving performance across all metrics. Similarly, for $s = 32$, decreasing $w$ from 128 to 64 slightly increases the FLOPs, from 102.92G to 105.66G, but results in significant performance gains across all metrics.
This behavior can be explained by the fact that smaller $w$ and $s$ values provide higher details of fMRI signals, allowing the model to process longer sequences for more detailed learning, albeit at the cost of higher computational complexity. These findings indicate that higher details of fMRI signals yield better results but require greater computational resources in terms of both FLOPs and parameters. These findings are also aligned with the scalability of the proposed method.}
\begin{table}[!ht]
\caption{Network size and computational complexity w.r.t different settings}
\label{tab:abl-complexity}
\resizebox{\textwidth}{!}{%
\begin{tabular}{|l|l|l|l|ll|l|l|}
\hline
\multirow{2}{*}{window ($w$)} & \multirow{2}{*}{stride ($s$)} & \multirow{2}{*}{GFLOPS} & \multirow{2}{*}{\#params} & \multicolumn{2}{l|}{COCO} & ADE20K & NSD \\ \cline{5-6}
 &  &  &  & \multicolumn{1}{l|}{AP$^\text{box}$} & {AP$^\text{segm}$} & mIoU & PCC \\ \hline
128 & 64 & 51.86 & 543.47M & \multicolumn{1}{l|}{41.9} & 41.1 & 40.90 & 55.80 \\
128 & 32 & 102.92 & 543.47M & \multicolumn{1}{l|}{42.3} & 41.6 & 41.80 & 56.20 \\
64 & 32 & 105.66 & 543.47M & \multicolumn{1}{l|}{\textbf{45.1}} &\textbf{ 44.0} & \textbf{42.92} & \textbf{57.43} \\ \hline
\end{tabular}%
}
\end{table}

\section{Conclusions and Discussions}
\label{sec:conclusion}
In this paper, we have investigated the feasibility of transferring human brain activities to the vision models via fMRI signals. In our proposed Brainformer, we introduce the concept of the fMRI feature technique that explores the local patterns of the signals. Additionally, the Brain 3D Voxel Embedding is proposed to preserve the 3D information that fMRI signals have been missed. The Multi-scale fMRI Transformer is presented to learn the features of a long signal. We also introduce Brain fMRI Guidance Loss inspired by the brain's mechanisms to transfer the vision semantic information of humans into the vision models. The empirical experiments on various benchmarks demonstrated that Brainformer is competitive with another SOTA method that uses text modality for knowledge transfer.
\newline
\noindent
\textbf{Limitations and Future Works}. 
Inspired by prior studies in neuroscience, we have developed our approach and demonstrated the efficiency of fMRI for training vision models. However, the proposed approach can potentially consist of minor limitations. 
First, although the NSD dataset is one of the large-scale datasets in the neuroscience field, compared to large-scale datasets such as ImageNet \cite{imagenet} in the computer vision field, this data is relatively small due to time and human efforts. However, we have demonstrated the performance of Brainformer with respect to the number of training data as in Section \ref{sec:abl_amount_data}. The result shows that if the amount of data is sufficient, the performance of Brainformer can be better and potentially surpass the previous methods. Second, the primary goal of this study is to present a new perspective on how to involve human recognition mechanisms in training vision learning models. The experimental configurations such as utilizing CLIP \cite{clip}, ConvNext \cite{liu2022convnet}, SwinTransformer \cite{liu2021swin} or even methods for downstream tasks, i.e., object detection, instance segmentation, semantic segmentation, and human brain response prediction, are conducted to prove our hypothesis on a fair basis. We leave further experiments with other settings for future studies.

\noindent
\textbf{Broader Impacts}.
Brainformer has a bi-directional impact on computer vision and the neuroscience field. In particular, we have illustrated how brain activities could help the vision model. Besides, Brainformer has potential benefits for neuroscientists to study brain activities, especially human cognition. Given a pair of images and corresponding fMRI, neuroscientists can utilize Brainformer as a valuable tool to explore which neurons inside the brain are highly activated with respect to the input image or particular objects inside, thus uncovering potential novel patterns of the brain.

\noindent
\textbf{Acknowledgements}

\bibliography{main, main_2}

\newpage

\begin{center}
    \section*{Author Biography}
\end{center}

\textbf{Xuan-Bac Nguyen} is currently a Ph.D. student at the Department of Computer Science and Computer Engineering of the University of Arkansas. He received his M.Sc. degree in Computer Science from the Electrical and Computer Engineering Department at Chonnam National University, South Korea, in 2020.  He received his B.Sc. degree in Electronics and Telecommunications from the University of Engineering and Technology, VNU, in 2015. In 2016, he was a software engineer in Yokohama, Japan. His research interests include Quantum Machine Learning, Face Recognition, Facial Expression, and Medical Image Processing.

\textbf{Xin Li} (Fellow, IEEE) received the BS degree with highest honors in electronic engineering and information science from University of Science and Technology of China, Hefei, in 1996, and the PhD degree in electrical engineering from Princeton University, Princeton, NJ, in 2000. He was a member of technical staff with Sharp Laboratories of America, Camas, WA from 2000 to 2002. He was a faculty member in Lane Department of Computer Science and Electrical Engineering, West Virginia University from 2003 to 2023. Currently, he is with the Department of Computer Science, University at Albany, Albany, NY 12222 USA. His research interests include image and video processing, compute vision and computational neuroscience. He was elected a Fellow of IEEE, in 2017 for his contributions to image interpolation, restoration and compression.

\textbf{Pawan Sinha} is a professor of vision and computational neuroscience in the Department of Brain and Cognitive Sciences at MIT. He received his undergraduate degree in computer science from the Indian Institute of Technology, New Delhi and his Masters and doctoral degrees in Artificial Intelligence from the Department of Computer Science at MIT. He has also had extended research stays at the University of California, Berkeley, Xerox Palo Alto Research Center, the Sarnoff Research Center in Princeton, and the Max-Planck Institute for Biological Cybernetics in Tübingen, Germany. Pawan’s research interests span neuroscience, artificial intelligence, machine learning, and public health. Using a combination of experimental and computational modeling techniques, research in his laboratory focuses on understanding the processes and principles of perceptual development. The lab's experimental work on these issues involves studying healthy individuals and also those with neurological disorders such as autism. The goal is not only to derive clues regarding the nature and development of human visual skills, but also to create more powerful and robust AI systems. Pawan has served on the program committees for prominent scientific conferences on object and face recognition and is currently a member of the editorial board of ACM's Journal of Applied Perception. He is a recipient of the Pisart Vision Award from the Lighthouse Guild International, the PECASE – the highest US Government award for young scientists, the Alfred P. Sloan Foundation Fellowship in Neuroscience, the John Merck Scholars Award for research on developmental disorders, the Jeptha and Emily Wade Award for creative research, the James McDonnell Scholar Award, the Troland Award from the National Academies, the Global Indus Technovator Award and the Distinguished Alumnus Award from IIT Delhi. Pawan's teaching has been recognized by Departmental honors and the Dean’s Award for Advising and Teaching at MIT. Pawan indulges, occasionally, in adventure sports, art and whimsical projects. He has trekked to the base of Mt. Everest, jumped out of a plane at 15,000 feet, contributed a regular comic strip to the MIT campus newspaper and been inducted into the Guinness Book of Records for creating the world’s smallest reproduction of a printed book. His first journey to the United States involved a plane crash and a ride aboard the Concorde.

\textbf{Samee U. Khan} received a Ph.D. in 2007 from the University of Texas, Arlington, TX. Currently, he is Professor of Electrical \& Computer Engineering at Mississippi State University (MSU) and served as Department Head from Aug. 2020 to July 2024. Before arriving at MSU, he was Cluster Lead (2016-2020) for Computer Systems Research at the National Science Foundation and the Walter B. Booth Professor at North Dakota State University. His research interests include optimization, robustness, and security of computer systems. His work has appeared in over 450 publications. He is the associate editor of IEEE Transactions on Cloud Computing and the Journal of Parallel and Distributed Computing. He is a Fellow of the IET and BCS. He is a Distinguished Member of the ACM and a Senior Member of the IEEE. 
He has won several awards, including the Best Paper Award (Systems Track), IEEE Cloud Summit, 2024; IEEE R3 Outstanding Engineer Award, 2024; IEEE Computer Society Distinguished Contributor Award, 2022 (inducted in the inaugural class); IEEE ComSoc Technical Committee on Big Data Best Journal Paper Award, 2019; IEEE-USA Professional Achievement Award, 2016; IEEE Golden Core Member Award, 2016; IEEE TCSC Award for Excellence in Scalable Computing Research (Middle Career Researcher), 2016; IEEE Computer Society Meritorious Service Certificate, 2016; Tapestry of Diverse Talents Award, North Dakota State University (NDSU), ND, USA, 2016; Exemplary Editor, IEEE Communications Surveys and Tutorials, IEEE Communications Society, 2014; Outstanding Summer Undergraduate Research Faculty Mentor Award, NDSU, ND, USA, 2013; Best Paper Award, IEEE Intl. Conf. on Scalable Computing and Communications (ScalCom), 2012; Sudhir Mehta Memorial International Faculty Award, NDSU, ND, USA, 2012; Best Paper Award, ACM/IEEE Intl. Conf. on Green Computing \& Communications (GreenCom), 2010.

\textbf{Khoa Luu} 
is an Assistant Professor and the Director of the Computer Vision and Image Understanding (CVIU) Lab in the Department of Electrical Engineering and Computer Science (EECS) at the University of Arkansas (UA), Fayetteville, US. He is affiliated with the NSF MonARK Quantum Foundry. He is an Associate Editor of the IEEE Access Journal and the Multimedia Tools and Applications Journal, Springer Nature. He is also the Area Chair in CVPR 2023, CVPR 2024, NeurIPS 2024, WACV 2025, and ICLR 2025. He was the Research Project Director at the Cylab Biometrics Center at Carnegie Mellon University (CMU), USA. His research interests focus on various topics, including Quantum Machine Learning, Biometrics, Smart Health, and Precision Agriculture. He has received eight patents and three Best Paper Awards and coauthored 120+ papers in conferences, technical reports, and journals. He was a Vice-Chair of the Montreal Chapter of the IEEE Systems, Man, and Cybernetics Society in Canada from September 2009 to March 2011. He was the Technical Program Chair at the IEEE GreenTech Conference 2024 and a co-organizer and chair of the CVPR Precognition Workshop in 2019-2024, MICCAI Workshop in 2019 and 2020, and ICCV Workshop in 2021. He was a PC member of AAAI, ICPRAI in 2020, 2021, 2022. He has been an active reviewer for several top-tier conferences and journals, such as CVPR, ICCV, ECCV, NeurIPS, ICLR, FG, BTAS, IEEE-TPAMI, IEEE-TIP, IEEE Access, Journal of Pattern Recognition, Journal of Image and Vision Computing, Journal of Signal Processing, and Journal of Intelligence Review.

\end{document}